\pdfoutput=1

\documentclass[11pt]{article}

\usepackage{acl}

\usepackage{times}
\usepackage{latexsym}

\usepackage{siunitx}

\usepackage[T1]{fontenc}

\usepackage[utf8]{inputenc}

\usepackage{microtype}

\usepackage{inconsolata}

\usepackage{graphicx}
\usepackage{lipsum}
\usepackage{dialogue}
\usepackage[inline]{enumitem}
\usepackage{subcaption}

\usepackage{authblk}

\usepackage{array}

\usepackage{tcolorbox}
\newcounter{promptno}[section]
\newlength\mystoreparindent
\newenvironment{prompt}[1][]
{
  \setlength{\mystoreparindent}{\the\parindent}
  \setlength{\parindent}{0pt}
  \refstepcounter{promptno}
  \par\medskip
  \noindent
  \begin{tcolorbox}[left=1pt,right=1pt]
  \textsc{{template \small\thesubsection.\thepromptno}}\\
  \small
  \tt
}{
  \end{tcolorbox}
  \setlength{\parindent}{\mystoreparindent}
  \medskip
}

\title{Retrieval-Augmented Code Generation for Situated\\ Action Generation: A Case Study on Minecraft}

\author[1]{Chalamalasetti Kranti}
\author[1]{Sherzod Hakimov}
\author[1,2]{David Schlangen}
\affil[1]{Computational Linguistics, Department of Linguistics \authorcr University of Potsdam, Germany}
\affil[2]{German Research Center for Artificial Intelligence (DFKI), Berlin, Germany}
\affil[ ]{\texttt{\{kranti.chalamalasetti, sherzod.hakimov, david.schlangen\}@uni-potsdam.de}}

\begin{document}
\maketitle
\begin{abstract}
In the Minecraft Collaborative Building Task, two players collaborate: an Architect (A) provides instructions to a Builder (B) to assemble a specified structure using 3D blocks. In this work, we investigate the use of large language models (LLMs) to predict the sequence of actions taken by the Builder. Leveraging LLMs' in-context learning abilities, we use few-shot prompting techniques, that significantly improve performance over baseline methods. Additionally, we present a detailed analysis of the gaps in performance for future work.

\end{abstract}

\section{Introduction}
Accurate action prediction is fundamental in developing interactive agents, enabling these systems to anticipate and effectively respond to user behaviors~\citep{DBLP:journals/ai/RoyR05, DBLP:journals/fthci/GoodrichS07}. Understanding language in dialogues is central to this process, as it involves interpreting instructions, intentions, and contextual cues to predict subsequent actions~\citep{DBLP:conf/aaai/ChenM11, DBLP:conf/iser/MatuszekHZF12, DBLP:conf/emnlp/Schlangen23}. %
Enhancing the precision of action prediction is particularly important in collaborative environments, where seamless interaction and coordination between agents and users are essential for achieving complex tasks~\citep{winograd1971procedures, DBLP:journals/ai/ThomazB08, DBLP:journals/arcras/TellexGKM20, DBLP:conf/corl/IchterBCFHHHIIJ22}.

\begin{figure}[ht]
  \includegraphics[width=0.5\textwidth]{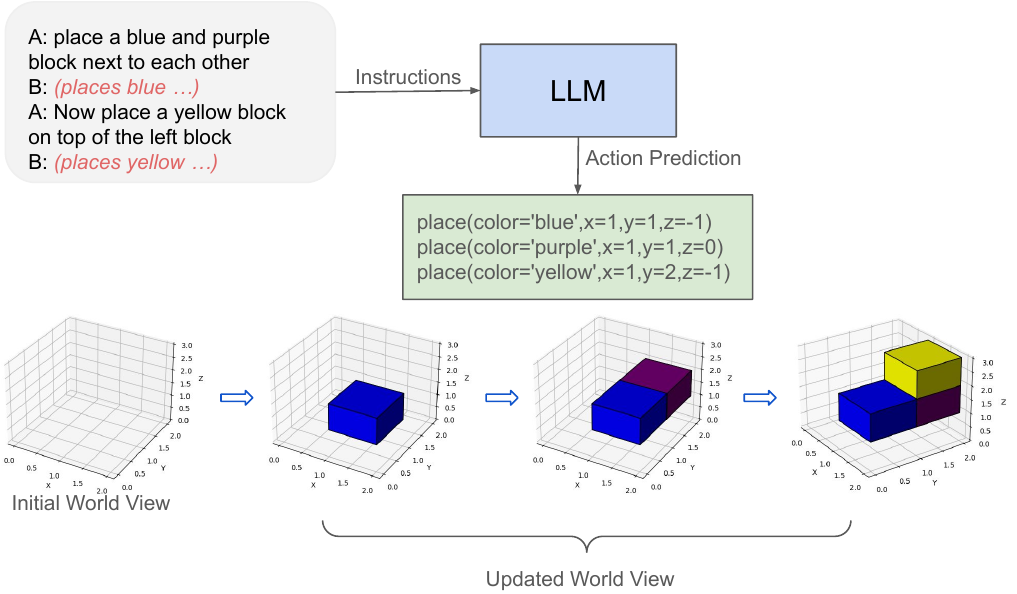}
  \caption{Illustration of the LLM interpreting block placement instructions. The initial world view is empty. The LLM receives instructions from User A and generates action predictions.}
  \label{fig:overview}
\end{figure}

Building on the importance of accurate action prediction and understanding language in dialogues, our research focuses on the Minecraft Collaborative Building Task~\citep{DBLP:conf/acl/Narayan-ChenJH19}. This task requires close coordination between an Architect (A) and a Builder (B). In this scenario, the Architect provides instructions to the Builder to assemble a specified structure using 3D blocks, making it an ideal testbed for investigating the capabilities of advanced computational models in interpreting and predicting actions based on natural language instructions.

Given the complexities involved in understanding natural language and predicting actions accurately, prior works~\citep{DBLP:conf/acl/JayannavarNH20, DBLP:conf/naacl/ShiFL22, DBLP:journals/corr/abs-2205-02388} proposed end-to-end neural models for modeling builder action prediction. However, achieving high accuracy for this task remains challenging. On the other hand, LLMs have demonstrated impressive performance in various natural language processing tasks, thanks to their ability to perform in-context learning~\citep{DBLP:conf/nips/BrownMRSKDNSSAA20}. Few-shot prompting techniques, in particular, allow these models to generalize from a limited number of examples~\citep{DBLP:journals/csur/LiuYFJHN23, DBLP:conf/nips/Wei0SBIXCLZ22, DBLP:journals/arobots/WuAKLZSBRF23}, making them well-suited for tasks requiring nuanced understanding and prediction of actions~\citep{DBLP:journals/corr/abs-2309-07864, DBLP:conf/icra/LiangHXXHIFZ23, DBLP:journals/arobots/SinghBMGXTFTG23, DBLP:conf/icml/DriessXSLCIWTVY23, DBLP:journals/access/VempralaBBK24}.

In this work, we explore the application of LLMs to predict the sequence of actions taken by the Builder (see Figure~\ref{fig:overview}). Taking advantage of the code-generation capabilities of LLMs, we model the action prediction task as a code-generation task.

\section{Related Work}
\textbf{Minecraft Collaborative Building Task:} \citet{DBLP:conf/acl/Narayan-ChenJH19} sourced human-to-human conversations for building structures in the Minecraft world. %
Building on this dataset~\citep{DBLP:conf/acl/JayannavarNH20} modeled builder action sequences using an encoder-decoder architecture %
with dialogue history and world state representation. %
Further advancing this task~\citep{DBLP:conf/naacl/ShiFL22} proposed a joint learning task for collaborative building and asking clarification questions. With a best F1-score of 0.21, this task remains unsolved.%

\paragraph{\textbf{IGLU - Multi Turn Dataset:}} \citet{DBLP:journals/corr/abs-2305-10783} features a human collaborating with an AI agent to build target structures, using 31 out of 150 original Minecraft building tasks. Since we focus on action generation for human-human interactions, this dataset falls out of scope for our evaluation.

\paragraph{\textbf{LLMs for Instruction Translation to a Code Snippet:}} Several research works~\citep{DBLP:journals/corr/abs-2107-03374, DBLP:conf/icml/HuangAPM22, DBLP:conf/iclr/ZengAICWWTPRSLV23, DBLP:journals/corr/abs-2303-18223} utilize LLMs for translating input natural language instructions to executable code snippets. These efforts span domains such as program synthesis~\citep{DBLP:conf/emnlp/0034WJH21, DBLP:journals/corr/abs-2302-13971, DBLP:journals/corr/abs-2308-12950, DBLP:journals/corr/abs-2403-00894} and robot instruction generation~\citep{DBLP:conf/icra/LiangHXXHIFZ23, DBLP:journals/arobots/SinghBMGXTFTG23, DBLP:conf/icml/DriessXSLCIWTVY23, DBLP:journals/access/VempralaBBK24, DBLP:journals/corr/abs-2404-09228}. This capability of LLMs motivates us to formulate the builder action prediction task as a code generation task.

\paragraph{\textbf{LLMs for Minecraft Collaborative Building Task}} \citet{DBLP:journals/corr/abs-2402-08392} %
use LLMs with text representation for action prediction, %
aligning closely with our proposal. However, %
we represent builder actions as code snippets and experiment on the Minecraft Dialogue dataset~\citep{DBLP:conf/acl/Narayan-ChenJH19}, %
providing a detailed analysis of action prediction performance.

\section{Dataset}
\label{sec:dataset}
Minecraft %
is used for studying collaborative tasks %
in a 3D voxel grid, %
with agents following architect's natural language instructions to build %
structures. The %
collaborative dialogue dataset~\citep{DBLP:conf/acl/Narayan-ChenJH19} %
has conversations for %
150 target structures with varying levels of abstractions and complexity, 
grouped into 547 dialogue games, %
split into train (309 games, 3,792 turn-code pairs), development (101 games, 1,335 turn-code pairs) and test sets (137 games, 1,615 turn-code pairs). %
We used the \textit{test set} for our evaluation.

\subsection{Builder Action Transformation}
Since the builder action prediction task is framed as a code-generation task, we convert the conversation from the format in the corpus (i.e., dialogue + action) into instruction + pseudo code. All the utterances of the builder and architect before each builder's action are aggregated into a single instruction. Builder actions involving \textit{puts down} are converted to \textit{place()}  function, and actions involving \textit{picks up} are converted to \textit{pick()} function. The following is a sample representation of this conversion.\footnote{%
    We will make the converted corpus available by the time of the conference.
}

\begin{dialogue}
\footnotesize
\label{shortdialogue}
\speak{Instruction} One block away from the edge, place a green block
\speak{Builder Action} Builder puts down a green block at X:0 Y:1 Z:4
\speak{\textbf{Code-Representation}} {\footnotesize\texttt{place(color=`green', x=0, y=1, z=4)}}
\newline
\vspace*{-.25\baselineskip}
\speak{Instruction} remove the middle block
\speak{Builder Action} Builder picks up a red block at X:0 Y:2 Z:0
\speak{\textbf{Code-Representation}} 
{\footnotesize\texttt{pick(color=`red', x=0, y=2, z=0)}}
\end{dialogue}

\section{Experimental Setup}
\label{sec:expsetup}
\paragraph{Few-shot Prompting} 
Following standard prompting approaches~\citep{DBLP:conf/nips/BrownMRSKDNSSAA20, DBLP:journals/csur/LiuYFJHN23, DBLP:conf/nips/Wei0SBIXCLZ22, DBLP:journals/arobots/WuAKLZSBRF23}, we adopt few-shot prompting to probe LLMs.
Our prompt (see Figure~\ref{fig:promptstructure} in Appendix) includes environment information such as the voxel grid size, available colors for the blocks and the quantity of each color. This is followed by dynamically adapted in-context examples suitable to the current turn instruction from the training set.

We use the pre-trained all-MiniLM-L6-v2 model from Sentence Transformers~\citep{DBLP:conf/emnlp/ReimersG19} to compute the similarity between the current turn instruction and all turns in the training set. The top three turns and their corresponding builder actions
are selected as in-context examples. For instance, if the current turn instruction is to ``start with a column of 5 purple bricks'', the in-context examples might be ``add two lines of purple bricks''
and ``start with a column of 5 red bricks''.
We conduct an ablation study on the impact of the parts of the prompt (refer to
section~\ref{subsec:ablationstudy} in Appendix) for the proposed task. Finally we append the test instruction and then query instruction-tuned %
LLMs such as \textit{GPT-4} (version o), \textit{Llama-3-8b} and \textit{Llama-3-70b} (see  Section~\ref{subsec:modelvariants} in Appendix for model selection).  All models are queried with temperature 0 and \texttt{max\_new\_tokens} = 500. 

\paragraph{LLM Fine Tuning}
In addition to few-shot prompting, we %
explored %
fine-tuning %
the Llama-3-8b model %
on the Minecraft building task %
to improve its generation of builder action sequences. We use the training set for the fine-tuning experiments. Additional details about the fine-tuning process are available Appendix~\ref{sec:appendix_finetuning}.

\paragraph{Evaluation metrics} We follow the same evaluation strategy as \citet{DBLP:conf/acl/JayannavarNH20}. %
Each dialogue turn prompts the LLM to generate action sequences (code) that provide information about a single block. The predicted sequences are compared against ground truth in the test split by checking block color, X, Y, Z coordinates, and command type (pick or place). This procedure is applied to all action sequences. We report the micro-averaged F1 on all dialogue turns.

\section{Results \& Analysis}
\label{sec:results}
The results %
compared to the baseline %
Builder Action Prediction (BAP) model %
by~\citet{DBLP:conf/acl/JayannavarNH20}, are presented in Table~\ref{tab:results}. GPT-4 achieved the best result (\num{0.39}) closely followed by Llama-3-70b (\num{0.33}). The fine-tuned version of Llama-3-8b showed a $\sim 6\%$ improvement over the vanilla version. Even though GPT-4 significantly %
outperforms the fine-tuned baseline, the upper bound for this task %
remains low. To understand this, %
we analysed %
the dialogues and identified references to spatial relations, real-world/geometric shapes, and anaphora. We then show how the GPT-4's performance in these categories. The performance of other models is discussed in Section~\ref{sec:appendix_erroranalysis} in the Appendix. Additionally, we identified two more factors complicating the interpretation of architect utterances, which may further impact action prediction.

\paragraph{\textbf{Spatial Prepositions:}} These are cases where the architect's utterances include prepositions that refer to a specific position in the grid, e.g. ``put another blue block on \textit{top} of it'' or ``two blocks on the ground, and one above on the \textit{left}''. %
Using Stanza~\citep{DBLP:conf/acl/QiZZBM20} and spaCy~\footnote{\url{https://spacy.io/}} %
we extract phrases %
with the Part-of-Speech (POS) tags: adverb (ADP), noun (NN), and preposition (IN), %
then manually filter out %
non-spatial phrases resulting in
\num{135} words. %
Examples include ``left'', ``right'', ``top'', ``bottom'', ``down'', ``front``, ``back'', ``towards'', ``between'', ``behind'', ``opposite'', ``parallel'', and ``inverse''. %
In the test set, \num{75.42}\% of the utterances include %
a spatial preposition,  %
but the model %
correctly generates code for only \num{26.03}\% of %
them, highlighting significant challenges in interpreting spatial prepositions and a need for improvement.

\begin{table}[]
\small
\centering
\begin{tabular}{lc}
\hline
\textbf{Model}   & \textbf{F1}   \\ \hline
GPT-4            & \textbf{0.39} \\
Llama-3-70b       & 0.33          \\
Llama-3-8b      & 0.18          \\
Llama-3-8b (fine-tuned)  & 0.19  \\ \hline
BAP (fine-tuned) & 0.21          \\ \hline
\end{tabular}%
\caption{Micro-Average F1-score for the builder action prediction task. The BAP (fine-tuned) model results are reported for H2 in game history and with 4x data augmentation~\citep{DBLP:conf/acl/JayannavarNH20}}
\label{tab:results}
\vspace*{-.5cm}
\end{table}

\paragraph{\textbf{Geometric and Real-World Shapes:}} %
Architect's utterances often include noun phrases referring to geometric shapes or real-world objects. %
Using the same procedure as above, we extracted 148 relevant nouns, %
including ``trident'', ``chair'', ``pitchfork'', ``circle'', and ``rectangle''. In the test set, \num{29.85}\% of utterances include one of %
these shapes, but the model %
correctly generated code for only \num{18.26}\% of them. %
This indicates that while LLMs have extensive general knowledge, they struggle with code generation for shape references, necessitating further examination to address these challenges.

\paragraph{\textbf{Anaphora:}} %
Architect's utterances often include pronouns %
referring to %
previous %
concepts, e.g. ``put another block next to \textit{it}''. %
Using the same procedure as above, we extracted and manually filtered these pronouns, identifying \num{16} words %
such as ``that'', ``this'', ``those'', ``it''. \num{46.81}\% of utterances in the test set include one of these pronouns %
but the model generated correct code only for \num{25.53}\% of %
them. Unlike spatial and geometric references, not all anaphora utterances indicate a reference to building concepts. They can also include acknowledgments or other types of information.

\begin{figure*}
\centering
  \includegraphics[scale=0.70]{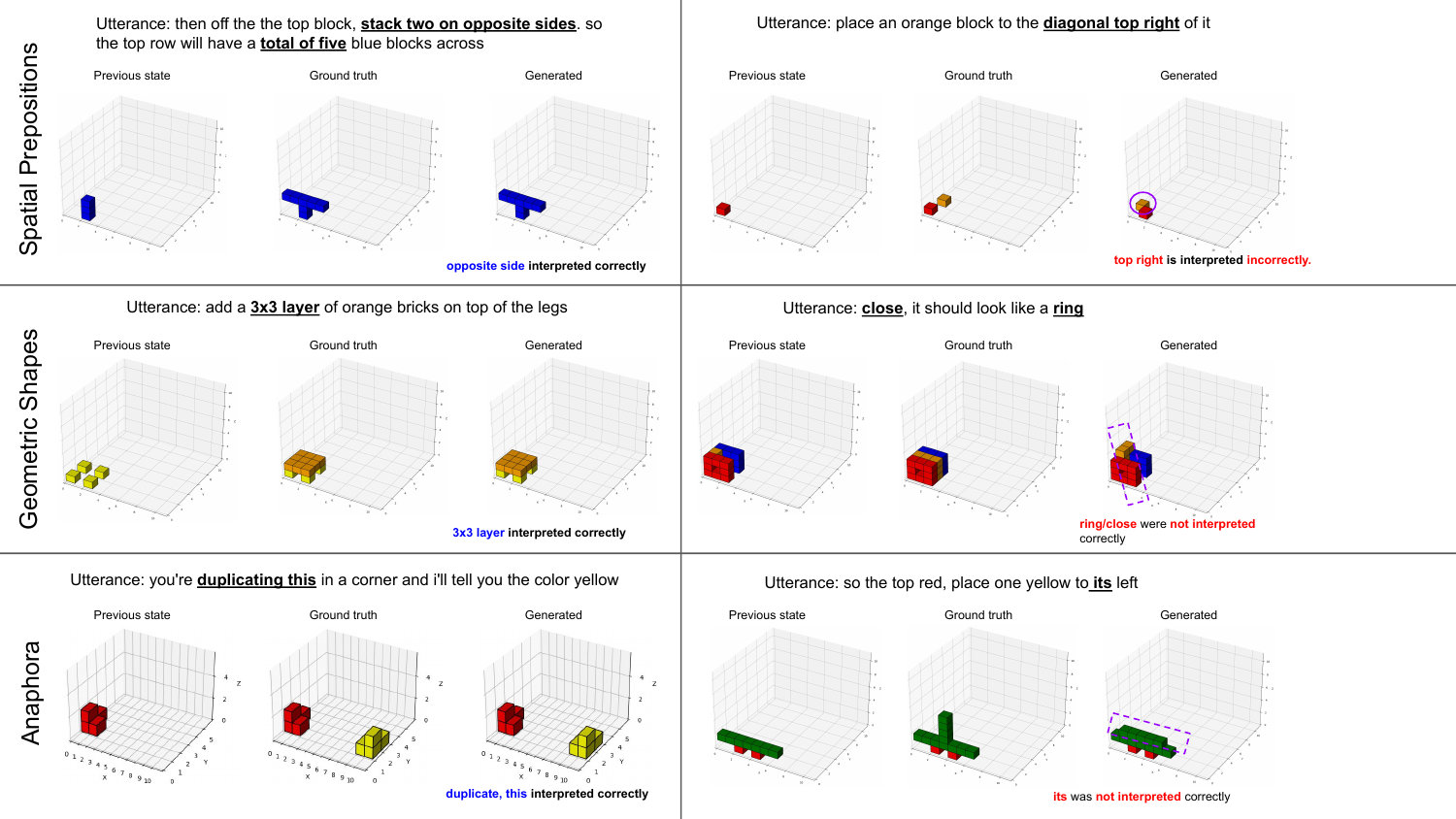}
  \caption{Voxel representations for sample turns that correspond to spatial preposition, geometric shape, and anaphora categories. Two samples for each category are given. Samples on the left side are generated correctly while samples on the right hand side have mistakes that are highlighted.}
  \label{fig:voxel_rep}
\end{figure*}

\paragraph{\textbf{Builder Mistakes:}} These %
occur when the builder %
places a block and later removes it, leaving %
spurious action sequences---one \textit{place} and one \textit{pick}---%
in the ground truth. %
E.g.``puts down a red block at X:-2 Y:0 Z: -1 followed by picks up a red block at X:-2 Y:0 Z:-1 (for a detailed example, please refer to Figure~\ref{fig:builder_mitsakes} in Appendix). %
Such cases cause evaluation mismatches. We filtered these by identifying \textit{place} commands followed by \textit{pick} commands at the same position and block color, finding that 23.3\% of turns contained such mistakes. These inaccuracies in the ground truth lead to lower evaluation scores, as the model is penalized for errors that are not indicative of its true performance.

\paragraph{\textbf{Underspecified Instructions:}} These are cases where the architect's utterance is underspecified, meaning it may lack specific details such as colour or precise location, or it may have multiple possible interpretations. 
To give an example, in the snippet shown in 
Figure~\ref{fig:underspecified_instructions} (in Appendix), there is no explicit indication in the dialogue history to interpret the instruction in a particular way, and the builder is left to decide on their own. Cases like this highlight the challenge of achieving a 100\% match with the ground truth.

\subsection{Qualitative Analysis} 
In Figure~\ref{fig:voxel_rep}, we provide qualitative examples for GPT-4 that illustrate the scenarios for spatial prepositions, geometric shapes, and anaphora. We have included two samples for each scenario: one where the model response aligns with the ground truth (left), and another where it does not (right). The model is able to handle spatial references such as ``opposite'', geometric shapes like ``3x3'', and repetitions like ``duplication''. However, it encounters challenges in accurately maintaining the exact count of blocks, translating words such as ``close'', ``ring'' into actions and resolving what ``it'' refers to.

\section{Conclusion}
We
investigate prompting large language models to build complex structures for the Minecraft building task. Predicting suitable builder action sequences for a given natural language instruction is challenging as the large language models need to accurately interpret language abstractions, decode spatial co-references and reason about repetitions solely based on the in-context samples in the prompt. We compared multiple instruction-tuned LLMs, both closed and open-source. In addition to showcasing improved performance over baseline results, we also conduct an in-depth analysis of the generated responses across multiple dimensions. In the future, we plan to investigate model architectures that can address the shortcomings identified in the evaluated models. 

\section*{Limitations}
Like all other prompting approaches, our approach needs to be more robust to the usability of pre-trained large language models.
\begin{enumerate*}[label=(\roman*)]
\item Not all grounded locations conform to the dimensions of the grid
\item Fails in interpreting instructions involving agent's perspective such as ``towards your right'', ``behind you'', ``away from you'' etc.
\item Struggles in understanding abstractions in the dialogue
\item Since the study is carried out in a simulated world, extending it to real-world agents may lead to incorrect consequences.
\end{enumerate*}.

\section*{Ethics Statement}
The research uses an open-source dataset (Minecraft dialogue corpus) and open-access, closed-API based pre-trained large language models. While these models are accessible, their usage is subject to legal restrictions as outlined in their respective terms of service and licensing agreements. Minecraft dialogue corpus does not include players’ personal, private information and does not contain any offensive conversations. However, the pre-trained LLMs, which inherit biases from their training data, may lead to code that favors certain styles and neglects others, hindering code portability. Another concern is the potential complexity of LLM-generated code, which can hinder end-user refinement and reuse. Moreover, it's crucial to ensure LLM-generated responses are free from harmful code, as their direct execution could impact the entire system.

\bibliography{custom}

\appendix

\section {Prompt Structure}
\label{sec:appendix_promptstructure}
The proposed task aims to investigate the capability of LLMs in accurately predicting actions based on input instructions. To generate an accurate action sequence, the models must understand the environment, follow dialogue history, and interpret the input instruction within the current context.

To achieve this, we construct a multi-part prompt, as illustrated in Figure~\ref{fig:promptstructure}, to query the LLM. This prompt includes several components: system information, environment information, task information, context information, and other guidelines. The system information specifies the expected behavior for the LLM. The environment information provides detailed information about the build region, its boundaries, and block properties. Specifically, the Minecraft dialogue corpus limits the build region to an 11x9x11 grid, and the builder is provided with an inventory of 120 blocks in six colors $colour \in$ (\textit{red}, \textit{blue}, \textit{orange}, \textit{purple}, \textit{yellow}, \textit{green}). The task information outlines the format of the expected outcome.

Following this, the context information includes in-context examples relevant to the current test instruction. Adopting the approach recommended by \citet{DBLP:conf/iccv/SongSWCW023}, we use adaptive prompts, dynamically retrieving examples from the training set that are similar to the current test instruction. Figure~\ref{fig:incontext-examples} shows how the in-context examples change based on the test instruction. This structure offers detailed information for the LLM to generate action sequences for the input instructions within the given environment.

\begin{figure*}
  \centering
    \begin{prompt}
\\
\textbf{System Info}
\\\\
You are an expert at interpreting natural language instructions and translating them into specific actions for placing colored blocks within a 3D grid.
\\\\
\textbf{Environment Info}
\\\\
The environment is structured as an 11x9x11 grid, with each cell representing a single block. The grid is organized such that columns align along the X-axis and rows along the Z-axis, employing Python indexing for cell identification. The X and Z axes range from -5 to 5, encompassing length and width, while the Y-axis ascends from 1 to 9, indicating height. This grid, following an XZY orientation, places X and Z as horizontal dimensions and Y as the vertical dimension. The ground level is defined at Y=1, with higher Y-values signifying increased elevation, thereby representing the vertical aspect of the environment. There are 20 blocks for each of the following colors: blue, orange, red, green, yellow, and purple.
\\\\
Please import the following functions for object interactions:
\\
pick(color,x,y,z): This function picks the object at the specified location.
\\
place(color,x,y,z): This function places the object at the specified location.
\\\\
You should only use these specified functions and not create any hypothetical functions.
\\\\
Please use the pick function only to select the pieces that you have previously placed on the board. You can safely assume that the available blocks are automatically selected, and therefore, you can directly use the "place()" function without needing to call the "pick()" function first.
\\\\
\textbf{Task Info}
\\\\
Your task is to interpret natural language instructions and generate the corresponding code representation. For each instruction labeled "Instruction" please respond with code under the label "Output" followed by a newline.
\\\\
\textbf{Context Info}
\\\\
\$INCONTEXT\_SAMPLES
\\\\
\textbf{Other Info}
\\\\
Do not generate any other text/explanations. Use python code to express the solution. Please keep the solutions simple and clear. Do not use loops, comments in the solution. Do not generate the instructions on your own. Stick to the given format. Think step by step.
\\\\
Let's get started.
\\\\
\$TEST\_INSTRUCTION
\end{prompt}
\caption{Prompt template used for the action prediction task. The system information specifies system level behavior, the environment information indicates the environment details of the user-agent environment, the context information describes the in-context examples, task information indicates the specific response format to follow.}
    \label{fig:promptstructure}
\end{figure*}

\begin{figure*}
  \centering

\begin{prompt}
\\
\textbf{INCONTEXT\_SAMPLES}
\\\\
The INCONTEXT\_SAMPLES varies for each turn. Below is an example
\\\\
\textbf{Instruction}
\\\\
Mission has started.. This structure looks fun- nothing will touch the ground!. cool!. We'll start with a row of 4 purple bricks, a layer above the ground
\\\\
\textbf{Output}
\\\\
place(color='purple',x=1,y=1,z=1)

place(color='purple',x=1,y=2,z=1)

pick(color='purple',x=1,y=1,z=1)

place(color='purple',x=0,y=2,z=1)

place(color='purple',x=-1,y=2,z=1)

place(color='purple',x=-2,y=2,z=1)
\\\\
\textbf{Instruction}
\\\\
Mission has started.. we'll start with two lines of purple bricks, parallel to one another. seperated by an empty space
\\\\
\textbf{Output}
\\\\
place(color='purple',x=1,y=1,z=-2)

pick(color='purple',x=1,y=1,z=-2)
\\\\
\textbf{Instruction}
\\\\
Mission has started.. We'll start with a column of 5 red bricks
\\\\
\textbf{Output}
\\\\
place(color='red',x=0,y=1,z=1)

place(color='red',x=0,y=2,z=1)

place(color='red',x=0,y=3,z=1)

place(color='red',x=0,y=4,z=1)

place(color='red',x=0,y=5,z=1)
\\\\

\textbf{TEST\_INSTRUCTION}
\\\\
Mission has started.. hi again!. Hello! This structure will start with a column of 5 purple bricks. It's all purple.

\end{prompt}
\caption{Retrieval of relevant in-context examples based on current test instruction}
    \label{fig:incontext-examples}
\end{figure*}

\subsection {Ablation Study}
\label{subsec:ablationstudy}
We investigate how the building blocks of the prompt structure shown in Figure~\ref{fig:promptstructure} impact overall task performance. Using the validation set of the Minecraft Dialogue dataset for our ablation study, we observe that the prompt structure with all components is optimal for the action prediction task and the LLMs in our experiments, as demonstrated in Table~\ref{tab:promptablationstudy}. Specifically, omitting in-context examples results in poor performance, while excluding environment information reduced the score slightly. Consequently, we use the prompt featuring all components with three in-context examples to ensure the best performance.

We also experimented with CodeLlama-34b~\footnote{\url{https://ai.meta.com/blog/code-llama-large-language-model-coding/}} model. However, the results were not sat-
isfactory because the generated response contained new instructions generated by model (hallucinations) and it was impossible to evaluate the closeness of the resulting generation to the ground truth, and hence this model is excluded from the experimental results.

\subsection {Model Variants}
\label{subsec:modelvariants}
For this study, we utilized three advanced large language models: GPT-4-O, Llama-3-8b, and Llama-3-70b. These models are instruction-tuned, enabling them to generate code snippets based on input text instructions. GPT-4-o and Llama-3-70b were accessed via their respective APIs, incurring a cost of \$12 for usage. For the Llama-3-8b model, experiments were conducted locally on a single A100 GPU with 80GB of memory, taking 10 hours in total.

\section {Fine tuning}
\label{sec:appendix_finetuning}
Since the out-of-the-box performance of the LLMs sets a lower bound for the action prediction task, we explore fine-tuning to see if it improves task performance. We use the training set of the Minecraft Dialogue corpus and fine-tune the Llama-3-8b model. This deliberate choice is made because it is the smallest open-source model we have used and has the lowest performance among all models. For the fine-tuning process, we use Q-LORA~\citep{DBLP:conf/nips/DettmersPHZ23} to reduce the memory footprint during fine-tuning by applying low-rank adaptations, allowing efficient fine-tuning on limited hardware.

We experimented with hyperparameter changes (number of epochs, learning rate) and finally chose the optimal performance setting, which includes 15 epochs and a learning rate of 0.0002. The model is configured using the Adam optimizer with a batch size of 32. The training process spans 15 epochs, with an early stopping condition based on evaluation loss to prevent over-fitting. Specifically, if the validation loss does not decrease for 5 consecutive epochs, the training is stopped. The validation set is used for evaluation during training to monitor the model's performance and guide the early stopping mechanism.

This fine-tuned model is then used for testing on the test set and is indicated as the Llama-3-8b (fine-tuned) model in Table~\ref{tab:results}. Compared to the baseline (pre-trained) model, the fine-tuned model shows an enhancement of $\sim 6\%$ in the F1-score. This improvement, although marginal, indicates a potential to improve the model's ability for the action prediction task.

\section {Detailed Error Analysis}
\label{sec:appendix_erroranalysis}
In this section, we delve into the performance of other models (Llama-3-8b and Llama-3-70b) across the specific categories of spatial relations, real-world/geometric shapes, and anaphora, which were identified as key factors influencing the overall performance. Out of \num{75.42}\% of the spatial preposition references, Llama-3-70b could predict correct action sequences for \num{22.66}\% of utterances. For utterances containing geometric and real-world shapes, the model correctly predicted \num{14.11}\% out of \num{29.85}\% of utterances. Anaphora resolution showed an accuracy of \num{20.5}\% correctly predicted out of \num{46.81}\% of utterances. Similarly, the Llama-3-8b model predicted \num{9.44}\% correctly for spatial prepositions, \num{6.22}\% for real-world and geometric shapes, and \num{10.05}\% for anaphora. These results highlight the challenges both models face in comprehending and predicting builder actions based on these linguistic elements.

\begin{table*}
\begin{tabular}{cc}
\hline
\textbf{Prompt}                   & \textbf{F1} \\ \hline
System Info + Env Info + Task Info + Context Info (Zero Samples) + Other Info  & 0.15        \\
System Info + Env Info + Task Info + Context Info (One Sample) + Other Info    & 0.17       \\
System Info + Env Info + Task Info + Context Info (Two Samples) + Other Info   & 0.18           \\
System Info + Env Info + Task Info + Context Info (Three Samples) + Other Info & 0.18          \\
System Info + Env Info + Task Info + Context Info (Four Samples) + Other Info & 0.18        \\
System Info + Env Info + Task Info + Context Info (Five Samples) + Other Info & 0.18            \\
Env Info + Task Info + Context Info (Three Samples) + Other Info     &  0.19           \\
System Info + Task Info + Context Info (Three Samples) + Other Info    & 0.17            \\
System Info + Env Info + Context Info (Three Samples) + Other Info     & 0.17            \\
System Info + Env Info + Context Info (Three Samples)    & 0.17           
\end{tabular}%
\caption{Ablation study for the number of in-context examples and components of the prompt structre on validation split of the Minecraft dataset using Llama-3-8b.}
\label{tab:promptablationstudy}
\end{table*}

\begin{figure*}
\centering
  \includegraphics[width=\textwidth]{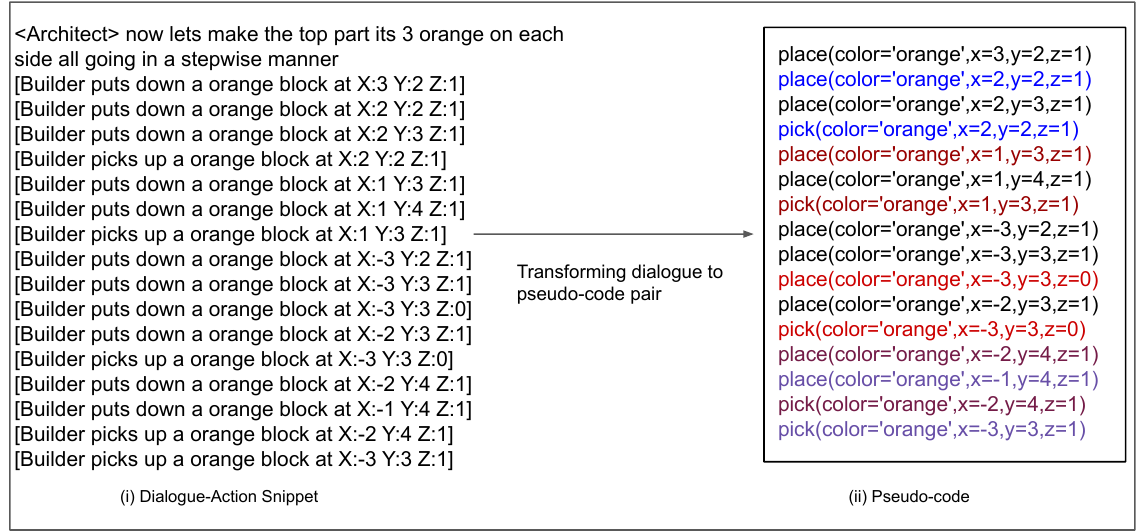}
  \caption{Excerpt of an utterance that contains the builder mistakes from the game-id: B29-A1-C151-1524078449685. The action sequence pairs where an item is first placed and later picked up is highlighted with the same colour.} 
  \label{fig:builder_mitsakes}
\end{figure*}  

\begin{figure*}
\centering
  \includegraphics[width=\textwidth]{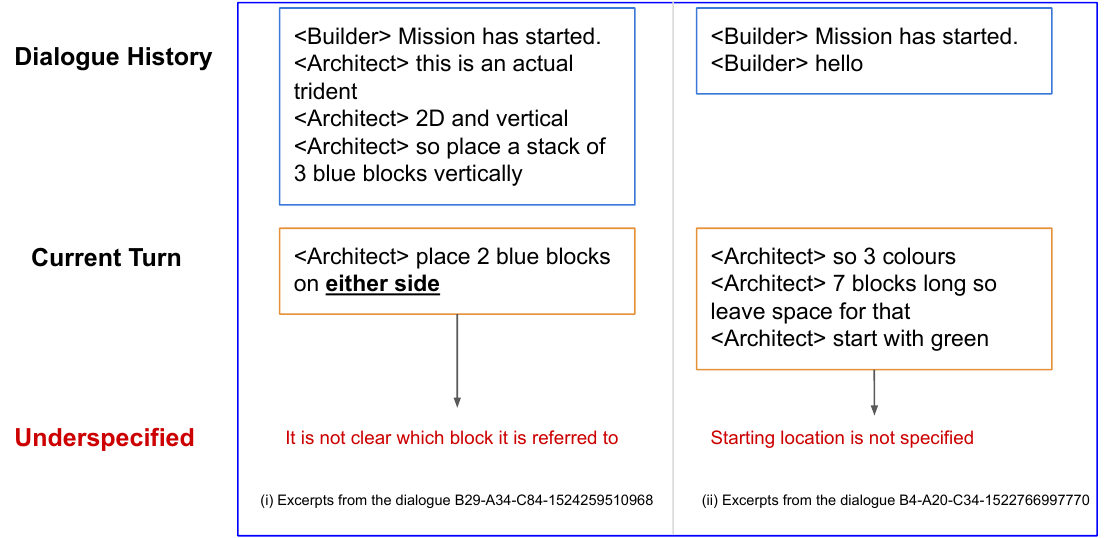}
  \caption{Examples utterances from the architect where the given command is underspecified} 
  \label{fig:underspecified_instructions}
\end{figure*}

\end{document}